# Expressive Completeness of Existential Rule Languages for Ontology-based Query Answering

**Heng Zhang**[1] and **Yan Zhang**[2] and **Jia-Huai You**[3]

[1]School of Computer Science & Technology, Huazhong University of Technology & Science, Wuhan, China
hengzhang@hust.edu.cn
[2]School of Computing, Engineering & Mathematics, Western Sydney University, Penrith, Australia
[3]Department of Computing Science, University of Alberta, Edmonton, Canada

## Abstract

Existential rules, also known as data dependencies in Databases, have been recently rediscovered as a promising family of languages for Ontology-based Query Answering. In this paper, we prove that disjunctive embedded dependencies exactly capture the class of recursively enumerable ontologies in Ontology-based Conjunctive Query Answering (OCQA). Our expressive completeness result does not rely on any built-in linear order on the database. To establish the expressive completeness, we introduce a novel semantic definition for OCQA ontologies. We also show that neither the class of disjunctive tuple-generating dependencies nor the class of embedded dependencies is expressively complete for recursively enumerable OCQA ontologies.

## 1 Introduction

Ontology-based Query Answering (OQA) has been a central issue in Ontological Reasoning and Knowledge Representation, and played an essential role in the Semantic Web, Data Modelling, and Data Exchange and Integration. Generally speaking, the problem of OQA may be stated as follows: Given a database $D$, an ontology (usually, a logical theory) $\Sigma$ and a query $q(\vec{x})$ with free variables $\vec{x}$, we need to find some answers $\vec{a}$ such that $D \cup \Sigma \models q(\vec{a})$ holds. In this setting, $D$ consists of a set of facts that have been observed, $\Sigma$ represents the domain knowledge, and $q$ describes the question about which we need to inquire. A major topic in OQA is to identify proper languages for expressing ontologies $\Sigma$.

Over the years, many languages have been developed for representing ontologies in OQA. In Description Logics, the DL-Lite family [Calvanese *et al.*, 2007], $\mathcal{EL}$ family [Baader *et al.*, 2005], and other variants have been proposed and extensively studied. More recently, existential rules, also known as data dependencies in Databases, and Datalog$^\pm$ in Knowledge Representation, have been rediscovered as a promising family of OQA languages, in which existential quantifiers, conjunctions and disjunctions may appear in the rule heads; see, e.g., [Baget *et al.*, 2011; Calì *et al.*, 2012; Alviano *et al.*, 2012; Bourhis *et al.*, 2013]. It is also worth noting some work on expressing ontologies by second-order logic, see, e.g., [Fagin *et al.*, 2005; Bourhis *et al.*, 2015].

While all these languages are of their specific features and hence are useful in different problem domains, some natural questions imposed to us are: Do we need to implement each of them? Can we have a universal language for OQA? On the other hand, as there are a large number of building blocks that have been introduced for specifying ontologies, it would be also interesting to know which of them is necessary for specifying ontologies in OQA. In this work, we try to answer these questions. In particular, we would like to know whether there is an expressively complete language for OQA.

Expressive completeness is one of the major principles for designing languages in Computer Science and Logic. For example, almost all the traditional programming languages, including Prolog, are known to be Turing complete; propositional logic can express all boolean functions; and by the well-known Lindström theorem, first-order logic is as expressive as any logic that enjoys both the compactness and the Löwenheim-Skolem property, see [Ebbinghaus *et al.*, 1994]. In Databases, the first-order language is shown to be complete for all the queries decidable in $AC^0$, and Datalog complete for the PTIME queries, see, e.g., [Immerman, 1999].

Very recently, by regarding existential rules as traditional database query languages, Gottlob *et al.* [2014b] showed that weakly-guarded existential rules capture the class of EXP-TIME queries, and Rudolph and Thomazo [2015] proved that tuple-generating dependencies capture the class of recursively enumerable queries. Both of these interesting characteraztions shed new insights on understanding the expressiveness of existential rules. However, it is also worth noting that, towards a full understanding of the expressiveness of OQA languages, treating them as traditional query languages is not enough. For example, many languages in the DL-Lite family are UCQ-rewritable, and almost all known tractable OQA languages are Datalog-rewritable; however, it is clearly not convenient to use UCQ and Datalog as OQA languages.

In this paper, we investigate the expressiveness of OQA languages in a more precise sense, and focus on Ontology-based Conjunctive Query Answering (OCQA), where conjunctive queries are used as the underlying query language. Our main contributions are as follows: Firstly, we propose a semantic framework for OCQA (Section 3). Such framework is general enough to unify different OCQA languages and hence can be used as a basis for studying the expressiveness of these languages for our purpose. In this framework, we then prove that disjunctive embedded dependencies (DEDs) precisely capture the class of recursively enumerable ontologies in OCQA (Section 4). Note that every set of DEDs can be rewritten as a set of disjunctive tuple-generating dependencies (DTGDs) and equality-generating dependencies (EGDs), two important classes of dependencies that have been extensively studied in OCQA. As an immediate consequence, our result implies a characterization of the expressiveness of DEDs as a traditional query language, which is similar to Rudolph and Thomazo's characterization [2015]. We also prove that neither DTGDs nor embedded dependencies (EDs) are expressively complete for recursively enumerable OCQA ontologies (Section 4). So the expressiveness between DEDs and DTGDs and EDs is explicitly differentiated. See [Zhang *et al.*, 2016] for a full version of this paper with more details.

## 2 Preliminaries

**Databases, instances and structures.** We assume $\Delta_c$ and $\Delta_v$ as infinite disjoint sets of *constants* and *variables*, respectively. *Terms* are constants or variables. A *relational schema* $\mathcal{R}$ is a set of *relation symbols*, and each of which is equipped with a natural number as its *arity*. Every *atom* over $\mathcal{R}$ is either an equality or a *relation atom* built upon terms and relation symbols in $\mathcal{R}$. A *fact* is a variable-free relation atom. Each *instance* over $\mathcal{R}$ is a set of facts over $\mathcal{R}$. Instances that are finite are called *databases*. For any instance $I$, let $adom(I)$ denote the set of constants occurring in $I$. Every *structure* $\boldsymbol{A}$ over $\mathcal{R}$ consists of a nonempty set $dom(\boldsymbol{A})$, named its *domain*, and an instance $I(\boldsymbol{A})$ over $\mathcal{R}$ such that $adom(I(\boldsymbol{A})) \subseteq dom(\boldsymbol{A})$. For convenience, sometimes an instance $I$ will be regarded as a structure $\boldsymbol{A}$ such that $dom(\boldsymbol{A}) = adom(I)$.

Let $\mathcal{R}$ be a relational schema, $\boldsymbol{A}$ and $\boldsymbol{B}$ be structures over $\mathcal{R}$, and $C$ be a subset of $dom(\boldsymbol{A}) \cap dom(\boldsymbol{B})$. A function $h : dom(\boldsymbol{A}) \to dom(\boldsymbol{B})$ is a *homomorphism* from $\boldsymbol{A}$ to $\boldsymbol{B}$ if, for all relation symbols $R \in \mathcal{R}$ and all constant tuples $\bar{a}$ of a proper length, $R(h(\bar{a})) \in I(\boldsymbol{B})$ if $R(\bar{a}) \in I(\boldsymbol{A})$. We write $h : \boldsymbol{A} \to_C \boldsymbol{B}$ if $h$ is a homomorphism from $\boldsymbol{A}$ to $\boldsymbol{B}$ such that $h(c) = c$ for all $c \in C$, and write $h : \boldsymbol{A} \Rightarrow_C \boldsymbol{B}$ if $h : \boldsymbol{A} \to_C \boldsymbol{B}$ and $h$ is injective. To simplify the presentation, $h$ will be dropped from the notations if it is clear from the context, and $C$ will be dropped if it is empty. Furthermore, if $h : \boldsymbol{A} \Rightarrow \boldsymbol{B}$ and an inverse $g$ of $h$ exists such that $g : \boldsymbol{B} \Rightarrow \boldsymbol{A}$, we call $h$ an *isomorphism* from $\boldsymbol{A}$ to $\boldsymbol{B}$, and write $h : \boldsymbol{A} \cong \boldsymbol{B}$. Given a relational schema $\mathcal{S} \subseteq \mathcal{R}$ and an instance $I$ over $\mathcal{R}$, let $I|_\mathcal{S}$ denote the set of facts involving relation symbols in $\mathcal{S}$ only.

**Queries and existential rules.** Let $\mathcal{R}$ be a relational schema. By a *query* over $\mathcal{R}$ we mean a first-order formula built upon atoms over $\mathcal{R}$ as usual. Given a query $\boldsymbol{q}$, let $const(\boldsymbol{q})$ denote the set of constants that occur in $\boldsymbol{q}$. A query is called *boolean* if it has no free variables. A *conjunctive query (CQ)* is a query of the form $\exists \vec{y}.\varphi(\vec{x}, \vec{y})$ where $\varphi$ is a conjunction of relation atoms. Let BCQ be short for boolean CQ. For each BCQ $\boldsymbol{q}$, we can regard it as a database by renaming existential variables as special constants and removing all quantifiers; let $[\boldsymbol{q}]$ denote such a database. Given a BCQ $\boldsymbol{q}$, the *graph* of $\boldsymbol{q}$ is an undirected graph with each atom in $\boldsymbol{q}$ as a node, and with each pair of atoms as an edge if they share an existential variable. A BCQ is called *prime* if its graph is connected.

Every *disjunctive embedded dependency (DED)* over $\mathcal{R}$ is a first-order sentence $\sigma$ of the following form:

$$\forall \vec{x} \forall \vec{y} (\phi(\vec{x}, \vec{y}) \to \exists \vec{z}_1.\psi_1(\vec{x}, \vec{z}_1) \lor \cdots \lor \exists \vec{z}_k.\psi_k(\vec{x}, \vec{z}_k)) \quad (1)$$

where $k \geq 1$, $\phi$ is a conjunction of relation atoms involving terms from $\vec{x} \cup \vec{y}$ only, each $\psi_i$ is a conjunction of atoms involving terms from $\vec{x} \cup \vec{z}_i$ only, and each variable in $\vec{x}$ has at least one occurrence in $\phi$. In particular, $\sigma$ is called an *embedded dependency (ED)* if $k = 1$, called a *disjunctive tuple-generating dependency (DTGD)* if it is equality-free, and called a *tuple-generating dependency (TGD)* if it is both an ED and a DTGD. The right-hand side of the implication is called the *body*, and left-hand side the *head*. Given a set $\Sigma$ of DEDs, a relation symbol in the relational schema of $\Sigma$ is called *intensional* if it occurs in the head of some DED in $\Sigma$, and *extensional* otherwise. For simplicity, we will omit the universal quantifiers in $\sigma$, and replace "$\wedge$" in the body by ",".

Let $D$ be a database over $\Sigma$ and $\boldsymbol{q}$ be a database, a set of DEDs and a BCQ, respectively, with a common relational schema $\mathcal{R}$. We write $D \cup \Sigma \models \boldsymbol{q}$ if, for all instances $I$ over $\mathcal{R}$ such that $D \subseteq I$, if $I \models \Sigma$ then $I \models \boldsymbol{q}$, where the satisfaction relation $\models$ is defined as usual. Let $\boldsymbol{p}(\vec{x})$ be a CQ over $\mathcal{R}$ with an $n$-tuple of free variables, the set of *answers of $\boldsymbol{p}$ w.r.t. $D$ under $\Sigma$*, denoted $Ans(D, \Sigma, \boldsymbol{p})$, is defined as the set of all $n$-tuples $\vec{a}$ of constants in $adom(D)$ such that $D \cup \Sigma \models \boldsymbol{p}(\vec{a})$.

**A modular lemma.** Below let us give a simple property that will be useful to verify the correctness of the OQA programming. We need to recall some notions first. An instance $I$ is called a *minimal model* of a set $\Sigma$ of DEDs if $I \models \Sigma$ and there exists no instance $J \subset I$ such that $J \models \Sigma$. Moreover, $\Sigma$ is *well-founded* if for every instance $J$ that satisfies $\Sigma$, there is a minimal model $I$ of $\Sigma$ such that $I \subseteq J$.

**Lemma 1** (Modular lemma). *Let $\boldsymbol{q}$ be a BCQ, and let $\Sigma$ and $\Gamma$ be two sets of DEDs such that $\Sigma$ is well-founded and possibly involves constants, no relation symbol from the relational schema of $\Sigma$ is intensional in $\Gamma$, and all relation symbols from the relational schema of $\boldsymbol{q}$ are intensional in $\Gamma$. Then $\Sigma \cup \Gamma \models \boldsymbol{q}$ iff $I \cup \Gamma \models \boldsymbol{q}$ for all minimal models $I$ of $\Sigma$.*

## 3 OCQA ontologies

As mentioned earlier, our task is to establish the expressive completeness of languages for OQA. To do this, we first have to answer the following questions firstly: What is an ontology in OQA? Can we have a general and semantic definition for OQA ontologies so that different languages can be unified in the same framework? Below we will address these issues.

### 3.1 Some general properties

Most languages that have been proposed for OQA are logical languages, including description logics, first-order logic, and sometimes fragments of higher-order logic. So, to study general properties that should be enjoyed by OQA ontologies, it is natural to consider those for ontologies as logical theories.

Instead of considering a specific logical language, we will regard every logical theory as a class of structures.

**Definition 1.** *Let $\mathcal{R}$ be a relational schema. An abstract theory over $\mathcal{R}$ is a class $\Phi$ of structures over $\mathcal{R}$ that is closed under isomorphisms, i.e., if $\boldsymbol{A} \cong \boldsymbol{B}$ and $\boldsymbol{A} \in \Phi$, then $\boldsymbol{B} \in \Phi$.*

It is clear that, given a concrete theory $\Sigma$ in any of the logical languages proposed to represent ontologies, the class of all models of $\Sigma$ is an abstract theory even if $\Sigma$ is infinite.

Let $\mathcal{R}$ be a relational schema. Let $D, \Phi$ and $\boldsymbol{q}(\vec{x})$ be a database, an abstract theory and a query over $\mathcal{R}$, respectively. Suppose $k$ is the length of $\vec{x}$. We let $Ans(D, \Phi, \boldsymbol{q})$ denote

$$\{\vec{a} \in adom(D)^k : \forall \boldsymbol{A} \in \Phi \, (\boldsymbol{A} \models D \Longrightarrow \boldsymbol{A} \models \boldsymbol{q}(\vec{a}))\}. \quad (2)$$

Slightly abusing the notation, we will not distinguish between abstract and concrete theories (e.g., TGD sets) if no confusion occurs. For instance, notations like $D \cup \Phi \models \boldsymbol{q}$ will be used.

The following proposition gives us some useful properties.

**Proposition 2.** *Let $\mathcal{R}$ be a relational schema; $\Phi$ be an abstract theory over $\mathcal{R}$; $D, D'$ be databases over $\mathcal{R}$; and $\boldsymbol{q}, \boldsymbol{q}'$ are boolean queries over $\mathcal{R}$. Then all the following hold:*

1. *If $D \cup \Phi \models \boldsymbol{q}$ and $D \cup \Phi \models \boldsymbol{q}'$ then $D \cup \Phi \models \boldsymbol{q} \wedge \boldsymbol{q}'$.*
2. *If $\boldsymbol{q} \models \boldsymbol{q}'$ and $D \cup \Phi \models \boldsymbol{q}$ then $D \cup \Phi \models \boldsymbol{q}'$.*
3. *If $D \Rightarrow_C D'$ and $D \cup \Phi \models \boldsymbol{q}$ then $D' \cup \Phi \models \boldsymbol{q}$, where $C$ denotes the set of constants occurring in $\boldsymbol{q}$.*

### 3.2 A semantic definition of OCQA ontology

With the properties we have presented, we are now in the position to give the semantic definition for ontologies in OQA. Note that our semantic definition is a refinement of the notion proposed by Arenas *et al.* [2014] where ontologies are regarded as arbitrary sets of database-query-answer triples.

**Definition 2.** Let $\mathcal{D}, \mathcal{Q}$ be relational schemas. Every *quasi-OCQA ontology* over $(\mathcal{D}, \mathcal{Q})$ is a set of pairs $(D, \boldsymbol{q})$ where $D$ is a database over $\mathcal{D}$ and $\boldsymbol{q}$ is a BCQ over $\mathcal{Q}$ with $const(\boldsymbol{q}) \subseteq adom(D)$. A quasi-OCQA ontology $O$ over $(\mathcal{D}, \mathcal{Q})$ is called an *OCQA ontology* if all the following conditions hold:

1. $O$ is *closed under query conjunctions*; that is, if both $(D, \boldsymbol{q}) \in O$ and $(D, \boldsymbol{q}') \in O$ then $(D, \boldsymbol{q} \wedge \boldsymbol{q}') \in O$.
2. $O$ is *closed under query implications*; that is, if $\boldsymbol{q} \models \boldsymbol{q}'$ and $(D, \boldsymbol{q}) \in O$ then $(D, \boldsymbol{q}') \in O$.
3. $O$ is *closed under injective database homomorphisms*; that is, if $D \Rightarrow_C D'$ and $(D, \boldsymbol{q}) \in O$ then $(D', \boldsymbol{q}) \in O$, where $C$ denotes the set of constants occurring in $\boldsymbol{q}$.

*Remark* 1. Quasi-OCQA ontologies defined above provide a rather relaxed way to define abstract OCQA ontologies. This notion is slightly more restricted than Arenas *et al.*'s notion (i.e., database-query-answer pairs, see [Arenas *et al.*, 2014]) as we require: (i) all the relational symbols involved in the database (resp., query) must belong to a given relational schema $\mathcal{D}$ (resp., $\mathcal{Q}$), and (ii) each constant involved in the query must appear in the corresponding database. Since $\mathcal{D}$ and $\mathcal{Q}$ could be the same, the first assumption actually does not lose any generality. In addition, as usual in Databases, it is also reasonable to assume that only constants that occur in the database will be involved in the query answering.

*Remark* 2. It is worth noting that the quasi-OCQA ontology is a very loose notion to define abstract OCQA ontologies. In fact, it is not hard to show that there are many quasi-OCQA ontologies that are not definable by any logical theory. This is the reason why we define the notion of OCQA ontologies.

**Definition 3.** Let $\mathcal{D}, \mathcal{Q}, \mathcal{R}$ be relational schemas such that $\mathcal{R} \supseteq \mathcal{D} \cup \mathcal{Q}$, and $\Phi$ be an abstract theory over $\mathcal{R}$. We define $Sem(\Phi, \mathcal{D}, \mathcal{Q})$ as the set of ordered pairs $(D, \boldsymbol{q})$ such that $D$ is a database over $\mathcal{D}$, $\boldsymbol{q}$ is a BCQ over $\mathcal{Q}$, and $D \cup \Phi \models \boldsymbol{q}$. Given an OCQA ontology $O$ over $(\mathcal{D}, \mathcal{Q})$, we say that $O$ is *defined by $\Phi$ over* $(\mathcal{D}, \mathcal{Q})$ if $O = Sem(\Phi, \mathcal{D}, \mathcal{Q})$.

With the above definition, any logical theory $\Phi$ can be related to abstract OCQA ontologies. By Proposition 2, we know that $Sem(\Phi, \mathcal{D}, \mathcal{Q})$ is always an OCQA ontology.

**Example 1.** Let $\Sigma$ be a set consisting of only the DED:

$$\bigwedge_{0 \leq i < 4} (\mathsf{A}(x_i, x_{s(i,4)}) \wedge \mathsf{A}(x_{s(i,4)}, x_i)) \to \mathsf{Goal} \vee \bigvee_{0 \leq i < j < 4} x_i = x_j \quad (3)$$

where $s(i, k)$ denotes $i + 1$ if $i < k$, and 0 otherwise. Let $D_k$ be the database $\{\mathsf{A}(a_i, a_{s(i,k)}) : 0 \leq i < k\} \cup \{\mathsf{A}(a_{s(i,k)}, a_i) : 0 \leq i < k\}$. It is a routine task to check that $D_4 \cup \Sigma \models \mathsf{Goal}$, and also easy to see that $D_3 \cup \Sigma \models \mathsf{Goal}$ does not hold.

Let $\mathcal{D} = \{\mathsf{A}\}$ and $\mathcal{Q} = \{\mathsf{Goal}\}$. Then $Sem(\Sigma, \mathcal{D}, \mathcal{Q})$ is an OCQA ontology which consists of pairs $(D, \mathsf{Goal})$ for all databases $D$ over $\mathcal{D}$ with a subset that is isomorphic to $D_4$.□

In Definition 3, only BCQs are considered. However, in query answering, non-boolean CQs will be used. The following fact shows that no generality will be lost in our setting.

**Proposition 3.** *Let $\mathcal{D}, \mathcal{Q}, \mathcal{R}$ be relational schemas such that $\mathcal{D} \cup \mathcal{Q} \subseteq \mathcal{R}$, and $\Phi, \Psi$ be abstract theories over $\mathcal{R}$ such that $Sem(\Phi, \mathcal{D}, \mathcal{Q}) = Sem(\Psi, \mathcal{D}, \mathcal{Q})$. Then for all databases $D$ over $\mathcal{D}$ and CQs $\boldsymbol{q}$ over $\mathcal{Q}$, $Ans(D, \Phi, \boldsymbol{q}) = Ans(D, \Psi, \boldsymbol{q})$.*

### 3.3 Expressive (in)completeness

Next, let us consider how to express arbitrary OCQA ontologies by DEDs. Due to a theoretical interest, we consider infinite sets of DEDs here. Such an investigation will provide a rational justification for our semantic definition.

**Proposition 4.** *Let $\mathcal{D}, \mathcal{Q}$ be two disjoint relational schemas, and $O$ be a quasi-OCQA ontology over $(\mathcal{D}, \mathcal{Q})$. Then $O$ can be defined by a possibly infinite set of DEDs iff it is an OCQA ontology.*

*Proof (Sketch).* The direction "only-if" is by Proposition 2. To show the converse, for each database-query pair from $O$, it is easy to construct a DED to generate its answers. □

In the above result, the DED set could be infinite. A natural question arises as whether all OCQA ontologies can be defined by a finite set of DEDs. For quasi-OCQA ontologies, the answer is negative. This is because there is some quasi-OCQA ontology which is highly undecidable, but query answering for finite sets of DEDs is recursively enumerable.[1] Naively, one might hope that the properties defining OCQA ontologies will get rid of the high undecidability. However, the following result tells us that this case will never happen.

**Proposition 5.** *There is some (highly undecidable) OCQA ontology that is not definable by any first-order sentence.*

*Proof (Sketch).* One can encode the recurring domino problem ($\Sigma_1^1$-hard [Harel, 1986]) by some OCQA ontology. Note that query answering with first-order sentences is in $\Sigma_1^0$. □

---

[1] Note that DEDs are first-order sentences and that the inference validity of first-order logic is recursively enumerable complete.

## 4 Recursively enumerable OCQA ontologies

In the last section, we have proved that the language consisting of finite sets of TGDs is not enough to capture the entire class of OCQA ontologies. This is not surprising because in the semantic definition of OCQA ontologies we do not consider whether the OCQA ontologies are realizable or not. In this section, we will focus on realizable OCQA ontologies, or more explicitly, recursively enumerable OCQA ontologies.

### 4.1 Machines that recognize OCQA ontologies

Before giving the characterizations, we first recall some notions and present a useful property. In the rest of this section, every *nondeterministic Turing machine (NTM)* is equipped with a read-only input tape and a read-write work tape. The head on the input tape will move right in all stages. Formally, an NTM $M$ is defined by a 6-tuple $(S, s_0, A, \Gamma, \Box, \delta)$, where

- $S$ is a finite and nonempty set of states, $s_0 \in S$ is the initial state, and $A \subseteq S$ is a set of accepting states;
- $\Gamma$ denotes the alphabet $\{0, 1, \Box, \sharp, \bar{0}, \bar{1}\}$,[2] where "$\Box$" will be used as the blank symbol;
- $\delta : (S \times \Gamma \times \Gamma) \to \mathcal{P}(S \times \Gamma \times \{L, R\})$ is a transition function, where $\mathcal{P}(X)$ denotes the power set of $X$ if $X$ is a set. Every *legal transition* of $M$ is a possible transition according to the transition function $\delta$ defined as usual.

Given an integer $k > 0$, we say that $M$ is *k-bounded* if for every triple $(s, a, b) \in S \times \Gamma \times \Gamma$ we have that $|\delta(s, a, b)| \leq k$. In particular, if $M$ is 1-bounded, we call $M$ *deterministic*, or directly say that $M$ is a *deterministic Turing machine (DTM)*. It is well-known that every NTM can be simulated by a DTM.

Given a natural number $k$, let $b(k)$ denote the binary representation (using "0" and "1") of $k$. We also fix $\mathcal{D}$ and $\mathcal{Q}$ as two relational schemas, $D$ as a database over $\mathcal{D}$, and $q$ as a BCQ over $\mathcal{Q}$. For $* \in \{d, q\}$, we let $w_*$ and $n_*$ denote the maximum arity and the number of relation symbols in $\mathcal{R}_*$, respectively. Let $c$ denote the size of $adom(D)$, and $e$ denote the number of existential quantified variables in $q$. Now we encode $(D, q)$ by the string, denoted by $[\![D, q]\!]$, as follows:

$$b(w_d) \sharp b(n_d) \sharp b(c) \sharp \overline{tvt(D)} \sharp b(w_q) \sharp b(n_q) \sharp b(e) \sharp tvt(q) \quad (4)$$

where $tvt(D)$ is a binary string encoding all truth values of ground atoms (built upon $adom(D)$ and $\mathcal{D}$) w.r.t. $D$; $tvt(q)$ is defined similarly by regarding $q$ as a special database; and for a given string $str$, let $\overline{str}$ denote the string obtained from $str$ by substituting $\bar{0}$ and $\bar{1}$ for 0 and 1, respectively.

**Example 2.** Suppose $\mathcal{D} = \{\mathsf{D}, \mathsf{E}\}$ and $\mathcal{Q} = \{\mathsf{Q}\}$, where $\mathsf{D}, \mathsf{E}$ are unary, and $\mathsf{Q}$ is binary. Suppose $D$ is a database consisting of a single fact $\mathsf{E}(a)$, and $q$ denotes the BCQ $\exists x \mathsf{Q}(x, a)$. Then we can encode $(D, q)$ by the string

$$1\sharp 10 \sharp 1 \sharp \bar{0}\bar{1} \sharp 10 \sharp 1 \sharp 1 \sharp 0010 \quad (5)$$

where by the string "$\bar{0}\bar{1}$" we mean that the fact $\mathsf{D}(a)$ is false in $D$, and the fact $\mathsf{E}(a)$ is true in $D$; by the string "0010" (in the last block) we mean that the atom $\mathsf{Q}(x, a)$ has an occurrence in $q$, while atoms $\mathsf{Q}(a, a), \mathsf{Q}(a, x)$ and $\mathsf{Q}(x, x)$ have not. □

---
[2] For a technical reason, we use two additional symbols "$\bar{0}$" and "$\bar{1}$" to represent special binary strings. It is easy to see that every NTM defined here can be simulated by one with a standard setting.

**Definition 4.** Let $\mathcal{D}$ and $\mathcal{Q}$ be relational schemas, and $O$ be an OCQA ontology over $(\mathcal{D}, \mathcal{Q})$. Then $O$ is *recognized* by an NTM $M$ if, for all databases $D$ over $\mathcal{D}$ and all BCQs $q$ over $\mathcal{Q}$, $(D, q) \in O$ iff $M$ accepts $[\![D, q]\!]$; $O$ is *recursively enumerable* if $O$ is recognized by some deterministic NTM.

To establish the desired expressive completeness, we need to simulate Turing machines by existential rules. The first obstacle to implement such a simulation is the lack of negation in the bodies of existential rules so that it would be difficult to represent the information "a fact is false in the database". Below we present a novel technique to address this issue.

Let $M$ be an abitrary NTM $(S, s_0, A, \Gamma, \Box, \delta)$. We call $M$ *convergent* if $\delta(s, \bar{0}, b) \subseteq \delta(s, \bar{1}, b)$ for all states $s \in S$ and all symbols $b \in \Gamma$. In other words, if $M$ is convergent, then every legal transition of $M$ on the input symbol $\bar{0}$ should be also a legal transition of $M$ on the input symbol $\bar{1}$.

Now let us explain the general idea of representing the negative information in a given database. Suppose $\mathsf{D}$ is a relation symbol in the database schema and $\tau$ is a transition of some convergent NTM $M$. Let $\phi_\tau(\vec{x})$ be a formula that describes the preconditions to implement the transition $\tau$ (except for the information about the truth of $\mathsf{D}(\vec{x})$), and $\psi_\tau(\vec{x})$ be a formula that defines the effect of implementing $\tau$. If $\mathsf{D}(\vec{x})$ is true in the database, one can simulate the transition $\tau$ by the rule

$$\mathsf{D}(\vec{x}), \phi_\tau(\vec{x}) \to \psi_\tau(\vec{x}). \quad (6)$$

For the case where $\mathsf{D}(\vec{x})$ is false in the database, by the encoding of the input we know that the bit which encodes the truth of $\mathsf{D}(\vec{x})$ is $\bar{0}$. Since $M$ is convergent, $\tau$ is legal for the case where $\mathsf{D}(\vec{x})$ is false implies that $\tau$ is legal for the case where $\mathsf{D}(\vec{x})$ is true. Thus we can now simulate $\tau$ by the rule

$$\phi_\tau(\vec{x}) \to \psi_\tau(\vec{x}). \quad (7)$$

According to the definition of the convergent NTM, it is easy to see that Rules (6) and (7) will always implement consistent actions. Note that no negation is involved in the simulation.

Now, let us present an interesting observation that will play an important role in establishing the main characterization.

**Proposition 6.** *Every recursively enumerable OCQA ontology can be recognized by a convergent, 2-bounded NTM.*

*Proof.* Let $O$ be a recursively enumerable OCQA ontology. Then there is a DTM, say $M$, that recognizes $O$. We assume $M = (S, s_0, A, \Gamma, \Box, \delta)$. Let $\delta_c$ be defined as follows:

$$\delta_c(s, a, b) = \begin{cases} \delta(s, \bar{1}, b) \cup \delta(s, \bar{0}, b) & \text{if } a = \bar{1}, \\ \delta(s, a, b) & \text{otherwise.} \end{cases} \quad (8)$$

Now let $M_c = (S, s_0, A, \Gamma, \Box, \delta_c)$. It is easy to see that $M_c$ is a convergent and 2-bounded NTM. So, it remains to show that $M_c$ also recognizes $O$. Given any input, it is clear that all accepting paths of $M$ are accepting paths of $M_c$. Thus, $M_c$ accepts all inputs $[\![D, q]\!]$ such that $(D, q) \in O$.

Conversely, let $(\mathcal{D}, \mathcal{Q})$ be the relational schema pair of $O$, $D$ be a database over $\mathcal{D}$, and $q$ be a BCQ over $\mathcal{Q}$ such that $M_c$ accepts the input $ins = [\![D, q]\!]$. We want to show that $(D, q) \in O$. Let $P$ be an accepting path that accepts $ins$. Let $ins_0$ be a string defined as follows: If $\bar{1}$ is scanned in the

$i$-th stage of $P$, and the next transition is not implemented by original actions in $\delta$, then let the $i$-th symbol of $ins_0$ be $\bar{0}$, otherwise let it be the same as that of $ins$. It is easy to see that $ins_0$ encodes a pair $(D', \boldsymbol{q})$ such that $D' \Rightarrow_C D$ where $C$ denotes $adom(D)$. Moreover, it is not hard to verify that $ins_0$ will be accepted by the original machine $M$, which means that $(D', \boldsymbol{q}) \in O$. Since $O$ is closed under injective database homomorphisms, we obtain that $(D, \boldsymbol{q}) \in O$ as desired. □

### 4.2 Expressive (in)completeness

In the following we present the main theorem, which states that the disjunctive embedded dependencies are expressively complete for recursively enumerable OCQA ontologies.

**Theorem 7.** *Let $\mathcal{D}, \mathcal{Q}$ be two disjoint relational schemas, and $O$ be a quasi-OCQA ontology over $(\mathcal{D}, \mathcal{Q})$. Then $O$ can be defined by a finite set of DEDs iff it is a recursively enumerable OCQA ontology.*

We leave the proof of this theorem to the next section.

Now, let us first present an interesting implication of our characterization. Some notions are needed. Fix $\mathcal{R}$ as a relational schema. A *database query* over $\mathcal{R}$ is a class of databases over $\mathcal{R}$ that is closed under isomorphisms. Similar to that in [Rudolph and Thomazo, 2015], let Goal be a special nullary relation symbol which does not belong to $\mathcal{R}$. A database query $Q$ is called *defined by* a set $\Sigma$ of DEDs if we have $D \in Q$ iff $D \cup \Sigma \models$ Goal. By Theorem 7, we then have a characterization for classical database queries, which is similar to Rudolph and Thomazo's characterization [2015]:

**Corollary 8.** *A database query $Q$ is defined by a finite DED set iff $Q$ is recursively enumerable and closed under injective homomorphisms, i.e., $D \in Q$ and $D \to D'$ implies $D' \in Q$.*

Next, let us consider the following question: Can the main theorem be established for a simpler language? For example, can equalities (resp., disjunctions) be removed so that DTGDs (resp., EDs) are still expressively complete? The following propositions show that, to assure the expressive completeness, neither disjunctions nor equalities can be removed.

Let us first define a notion. An OCQA ontology $O$ over a relational schema pair $(\mathcal{D}, \mathcal{Q})$ is called *first-order rewritable* if for all BCQ $\boldsymbol{q}$ over $\mathcal{Q}$ there is a first-order sentence $\boldsymbol{q}_O$ such that $(D, \boldsymbol{q}) \in O$ iff $D \models \boldsymbol{q}_O$ for all databases $D$ over $\mathcal{Q}$. With this notion, we are then in the position to present the results.

**Proposition 9.** *There is a first-order rewritable OCQA ontology that cannot be defined by any finite set of DTGDs.*

**Proposition 10.** *There is a first-order rewritable OCQA ontology that cannot be defined by any finite set of EDs.*

Proposition 9 can be proved by showing that every OCQA ontology defined by DTGDs is closed under database homomorphisms. Example 1 gives a first-order rewritable OCQA ontology that is not closed under database homomorphisms. The next proposition can be proved by showing that the complement of every OCQA ontology defined by EDs is closed under direct products; however first-order rewritable OCQA ontologies that do not have this property can be easily found.

*Remark* 3. The above incompleteness results show why disjunctions and equalities should be considered in identifying new first-order rewritable and tractable OCQA languages.

## 5 Proof of the main theorem

From now on, let us focus on the proof of Theorem 7. The direction "only-if" follows from Proposition 2 and the well-known fact that the inference validity of first-order logic is recursively enumerable complete. So, in this section we only consider the converse, which will be proved by constructing a finite set of DEDs. We first explain the general idea of the construction, and then implement it step by step.

### 5.1 General idea of the proof

In Databases, to show a language is expressively complete for a class of database queries in a certain level of complexity, a usual way is by constructing logical theories to simulate Turing machines, see, e.g., [Vardi, 1982; Rudolph and Thomazo, 2015]. However, it should be noted that establishing the expressive completeness of OQA languages is significantly different from that of classical querying languages.

The main difficulty is as follows: Although the BCQ is one part of the input for the query answering, the logical theory that simulates the query answering actually cannot access the BCQ, which means that, instead of focusing on one decision problem, we have to treat an infinite number of decision problems. Note that there are an infinite number of BCQs.

Fortunately, we have found a novel technique to overcome these obstacles. To explain it, we first define some notations, and then present a property. Let $\mathcal{R}$ be a relational schema, $\mathscr{D}$ be a class of databases over $\mathcal{R}$, and $C$ be a set of constants. We let $\bigoplus_C \mathscr{D}$ denote the instance $\bigcup\{D^* : D \in \mathscr{D}\}$ where, for every $D \in \mathscr{D}$, $D^*$ is an isomorphic copy (a database over $\mathcal{R}$) of $D$ such that, for any pair of distinct databases $D_1, D_2$ in $\mathscr{D}$, only constants from $C$ will be shared by $D_1^*$ and $D_2^*$.

The following is a key lemma to our proof.

**Proposition 11.** *Let $\mathcal{D}, \mathcal{Q}$ be relational schemas, and $O$ be an OCQA ontology over $(\mathcal{D}, \mathcal{Q})$. Then for all databases $D$ over $\mathcal{D}$ and BCQs $\boldsymbol{q}$ over $\mathcal{Q}$, $(D, \boldsymbol{q}) \in O$ iff $U_O(D) \models \boldsymbol{q}$, where $U_O(D) = \bigoplus_C \{[\boldsymbol{q}'] : (D, \boldsymbol{q}') \in O\}$ and $C = adom(D)$.*

*Proof.* The direction "only-if" is trivial; so we only show the converse. Let $\boldsymbol{q}$ be a BCQ over $\mathcal{Q}$ and $D$ be a database over $\mathcal{D}$ such that $U_O(D) \models \boldsymbol{q}$. By definition, there exists a homomorphism $h$ from $[\boldsymbol{q}]$ to $U_O(D)$ such that $h(c) = c$ for all constants $c$ from $C$. Let $\boldsymbol{q}'$ be a BCQ obtained from $\boldsymbol{q}$ by substituting $h(x)$ for all variables $x$ such that $h(x) \in C$, and let $h'$ be a restriction of $h$ to all variables and constants occurring in $\boldsymbol{q}'$. Then it is clear that $\boldsymbol{q}' \models \boldsymbol{q}$ and that $h'$ is a homomorphism from $[\boldsymbol{q}']$ to $U_O(D)$. Without loss of generality, let us write $\boldsymbol{q}'$ in the form $\boldsymbol{q}_1' \wedge \cdots \wedge \boldsymbol{q}_k'$, where $k \geq 1$ and each $\boldsymbol{q}_i'$ is prime (see Section 2 for the definition). Take $i \in \{1, \ldots, k\}$. Obviously, $h'$ is a homomorphism from $[\boldsymbol{q}_i']$ to $U_O(D)$. Since $\boldsymbol{q}_i'$ is prime, by the definition of $U_O(D)$ we know that there is at least one BCQ $\boldsymbol{p}_i$ such that $(D, \boldsymbol{p}_i) \in O$ and that $h'(\boldsymbol{q}_i')$ is contained in the disjoint copy of $\boldsymbol{p}_i$ in $U_O(D)$. From the latter we conclude that $\boldsymbol{p}_i \models \boldsymbol{q}_i'$. Since $O$ is closed under query implications, we infer that $(D, \boldsymbol{q}_i') \in O$. Furthermore, since $O$ is closed under query conjunctions, we obtain that $(D, \boldsymbol{q}') \in O$. As we have proved previously, it holds that $\boldsymbol{q}' \models \boldsymbol{q}$. Again by the query implication closure of $O$, we conclude that $(D, \boldsymbol{q}) \in O$, which completes the proof. □

With the property above, we call $U_O(D)$ a *universal model* of $O$ w.r.t. $D$. Now, let us explain the general idea of our proof. From now on, let us fix $\mathcal{D}$ and $\mathcal{Q}$ as two disjoint relational schemas, and $O$ as a recursively enumerable OCQA ontology over $(\mathcal{D}, \mathcal{Q})$. Our task is to construct a finite set $\Sigma$ of DEDs such that, for any given database $D$ over $\mathcal{D}$, $U_O(D)$ is isomorphic to the minimal model of $D \cup \Sigma$.

Now, we show how to construct the set $\Sigma$ of DEDs. Suppose $M$ is a Turing machine which recognizes $O$. For all possible BCQs $\boldsymbol{q}$ over $\mathcal{Q}$, we let $\Sigma$ work as follows:

1. Simulate the computation of $M$ on input $ins = [\![D, \boldsymbol{q}]\!]$.
2. If $M$ accepts $ins$ then copy $[\boldsymbol{q}]$ to the intended model.

To implement the simulation of $M$, we have to define natural numbers on the intended domain, which can be done in DEDs easily. Another difficulty is to encode negative information in the database as existential rules have no negations (see [Rudolph and Thomazo, 2015]). Thanks for the power of Proposition 6, this again can be overcome in a natural way.

### 5.2 Defining numbers and arithmetic relations

To define numbers and arithmetic relations, we have to generate a successor relation first. In particular, we require that the successor relation will travel over all the constants from the database so that every fact in the database could have a fixed encoding. Let DC be a fresh unary relation symbol that collects all the constants occurring in the intended database, which can be defined by the following DEDs:

$$\mathsf{D}_i(x_1, \ldots, x_k) \to \mathsf{DC}(x_1) \wedge \cdots \wedge \mathsf{DC}(x_k) \quad (9)$$

for all relation symbols $\mathsf{D}_i$ from the schema $\mathcal{D}$, where $k$ denotes the arity of $\mathsf{D}_i$. Let $\Sigma_s$ denote the set that consists of all the DEDs which define DC and all the following DEDs:

$$\mathsf{DC}(x), \mathsf{DC}(y) \to \mathsf{LT}(x,y) \vee x = y \vee \mathsf{LT}(y,x) \quad (10)$$
$$\mathsf{LT}(x,y), \mathsf{LT}(y,z) \to \mathsf{LT}(x,z) \quad (11)$$
$$\mathsf{LT}(x,x) \to \mathsf{Undesired} \quad (12)$$
$$\mathsf{DC}(x), \mathsf{DC}(y), \mathsf{LT}(x,y) \to \mathsf{NotMin}(y) \quad (13)$$
$$\mathsf{DC}(x) \to \mathsf{Min}(x) \vee \mathsf{NotMin}(x) \quad (14)$$
$$\mathsf{DC}(x), \mathsf{DC}(y), \mathsf{LT}(x,y), \mathsf{LT}(y,z) \to \mathsf{LTNotSucc}(x,z) \quad (15)$$
$$\mathsf{DC}(x), \mathsf{DC}(y), \mathsf{LT}(x,y) \to \mathsf{Succ}(x,y) \vee \mathsf{LTNotSucc}(x,y) \quad (16)$$
$$\mathsf{Succ}(x,y) \to \exists z. \mathsf{Succ}(y,z) \wedge \mathsf{LT}(y,z) \quad (17)$$
$$\mathsf{Succ}(x,y) \to \mathsf{Num}(x) \wedge \mathsf{Num}(y) \quad (18)$$

where $\mathsf{Succ}(x, y)$ asserts that $y$ is an immediate successor of $x$, and $\mathsf{LT}(x, y)$ states that $x$ is strictly less than $y$. In this way, a linear order would be arbitrarily generated by disjunctions in DED (10) and the existential quantifier in DED (17). Note that the linear order generated by such disjunctions and existential quantifier is not always good – sometimes a cycle could be there. In this case, we simply set the nullary relation Undesired true. As we will see later, this flag will lead to giving up the intended instance under the current focus.

If the flag Undesired is false, by the transitivity and irreflexivity, we know that LT is a linear order. Thus, we have:

**Proposition 12.** *Let $D$ be a database over $\mathcal{D}$. Then $D \cup \Sigma_s$ is well-founded, and for each minimal model $I$ of $D \cup \Sigma_s$ with Undesired $\notin I$, Succ defines an infinite successor relation on $adom(I)$, Min defines the minimum element w.r.t. Succ, and Num defines the set of all elements from $adom(I)$.*

With the above relations, we can define relations such as Add, Mul, $\mathsf{Bit}_b$ and Len, where $b \in \{0, 1\}$, Add$(x, y, z)$ (Mul$(x, y, z)$, respectively) asserts that $z$ is the sum (product, respectively) of $x$ and $y$, $\mathsf{Bit}_b(x, y)$ asserts that the $y$-th bit of the binary representation of $x$ is $b$. All of these can be defined in a routine way. For example, the addition relation can be defined by the following DEDs:

$$\mathsf{Num}(x), \mathsf{Min}(y) \to \mathsf{Add}(x, y, x) \quad (19)$$
$$\mathsf{Add}(x, y, z), \mathsf{Succ}(y, u), \mathsf{Succ}(z, v) \to \mathsf{Add}(x, u, v) \quad (20)$$

Finally, we let $\Sigma_{num}$ denote the union of $\Sigma_s$ and the set of all DEDs that define the mentioned arithmetic relations.

### 5.3 Simulating Turing machine

Now, we are ready to define some DEDs to simulate the Turing machine. According to Proposition 6, we only need to simulate a convergent, 2-bounded NTM $M$. As mentioned previously, we will simulate the computations of $M$ on all possible input queries in parallel. Towards this end, we introduce a number of relation symbols $\mathsf{ITape}_a$, $\mathsf{WTape}_a$, $\mathsf{IHead}$, $\mathsf{WHead}$ and $\mathsf{State}_s$, where $a$ is a symbol in the alphabet, and $s$ is a state. For example, $\mathsf{ITape}_a(x, y)$ asserts that, in the computation for the BCQ encoded by $y$, the symbol written on the input tape at position $x$ is $a$; $\mathsf{State}_s(x, y, z)$ asserts that, in the computation path (encoded by) $y$ for BCQ (encoded by) $z$, the state of $M$ in time $x$ is $s$. Note that, as $M$ is 2-bounded, the (nondeterministic) choices of all stages can be represented by a binary string in an obvious way.

The simulation of $M$ consists of three parts as follows.

**Initialization.** This part includes specifying the initial state, copying the string encoding a database-query pair to the input tape, and filling unused cells with the blank symbol. We only focus on the second one here. The other two are trivial.

Suppose relation symbols allowed in database (resp., BCQ) list as $\mathsf{D}_0, \ldots, \mathsf{D}_{m-1}$ (resp., $\mathsf{Q}_0, \ldots, \mathsf{Q}_{n-1}$). We first introduce some fresh relation symbols $\mathsf{PosD}_i$ and $\mathsf{PosQ}_j$, where $0 \le i < m$, $0 \le j < n$, $\mathsf{PosD}_i$ (resp., $\mathsf{PosQ}_j$) is of arity $k + 1$ if $\mathsf{D}_i$ (resp., $\mathsf{Q}_j$) is $k$-ary. Informally, $\mathsf{PosD}_i(\vec{x}, y)$ asserts that the truth value should be copied to the $y$-th cell of the input tape; the meaning of $\mathsf{PosQ}_j$ is defined similarly. Since it is a folklore in the KR community as how to define such relations by arithmetic relations, we omit the details here.

The DEDs below are designed to copy the database:

$$\mathsf{D}_i(\vec{y}), \mathsf{PosD}_i(\vec{y}, z), \mathsf{BCQ}(x) \to \mathsf{ITape}_{\bar{1}}(z, x) \quad (21)$$
$$\mathsf{PosD}_i(\vec{y}, z), \mathsf{BCQ}(x) \to \mathsf{ITape}_{\bar{0}}(z, x) \quad (22)$$

At a first glance, the above encodings seem problematic since $\mathsf{ITape}_{\bar{0}}(z, x)$ and $\mathsf{ITape}_{\bar{1}}(z, x)$ might be true simultaneously. However, by the convergence of $M$ and the analysis in Subsection 4.1 we know that this does not lead to any trouble.

Next, let us fix the way to encode BCQs (e.g., the Gödel numbering). By the arithmetic relations, we can define relations $\mathsf{BCQ}$, $\mathsf{HasQ}_i$ and $\mathsf{NHasQ}_i$ for all proper indices $i$ such that $\mathsf{BCQ}(x)$ asserts that $x$ encodes some BCQ over $\mathcal{Q}$, and $\mathsf{HasQ}_i(\vec{x}, y)$ (resp., $\mathsf{NHasQ}_i$) asserts that $\mathsf{Q}_i(\vec{x})$ appears (resp., does not appear) in the BCQ encoded by $y$. To copy the focused BCQ, we use the following DEDs:

$$\mathsf{HasQ}_i(\vec{y}, x), \mathsf{PosQ}_i(\vec{y}, z), \mathsf{BCQ}(x) \to \mathsf{ITape}_1(z, x) \quad (23)$$
$$\mathsf{NHasQ}_i(\vec{y}, x), \mathsf{PosQ}_i(\vec{y}, z), \mathsf{BCQ}(x) \to \mathsf{ITape}_0(z, x) \quad (24)$$

**Transition.** Simulating transitions of $M$ is almost the same as that in the classical case, see, e.g., [Dantsin *et al.*, 2001; Baget *et al.*, 2011]. The only difference here is that we have to treat nondeterminism. It should be noted that this cannot be directly encoded by disjunctions. However, we can first guess a number which encodes the action choices in all transitions, and then simulate the computation on these choices, which is deterministic and can be encoded in a routine way. Note that the "guess" will be implemented in the next part.

To carry out the idea, as mentioned in the begining of this subsection, for each relation contributing to defining the computation path, we use an additional argument to specify (the number encoding) the choices. Without loss of generality, let $\delta(s,c) = \{(s_0, c_0, d_0), (s_1, c_1, d_1)\}$ be a transition of $M$. Let $\phi_i(x, \vec{y}) \to \psi_i(\vec{z})$ be a DED to simulate the deterministic transition $\delta(s,c) = (s_i, c_i, d_i)$ in the $x$-th stage of the computation. Now, our transition can be simulated by

$$\mathsf{Bit}_i(v,x), \phi_i'(x, \vec{y}, v) \to \psi_i'(\vec{z}, v) \qquad (25)$$

for all $i \in \{0, 1\}$, where $\phi_i'$ and $\psi_i'$ are obtained from $\phi_i$ and $\psi_i$, respectively, by adding the argument $v$ to related atoms. Intuitively, these DEDs assert that $M$ will perform the first action if the $x$-bit of $v$ is 0, otherwise $M$ will carry out the second one.

**Acceptance.** Since $M$ is nondeterministic, to describe that $M$ accepts an input, we have to state that there is an accepting state that can be reached from the initial state via a sequence of action choices (encoded by a number). To implement this, for each accepting state $s$, we construct a DED as follows:

$$\mathsf{BCQ}(z), \mathsf{State}_s(x, y, z) \to \mathsf{Accept}(z) \qquad (26)$$

Finally, let $\Sigma_{sim}$ consist of all DEDs defined in this subsection. Then we have an encoding to simulate machine $M$.

**Proposition 13.** *Let $D$ be a database over $\mathcal{D}$, and $I$ be a minimal model of $D \cup \Sigma_{num} \cup \Sigma_{sim}$ such that $\mathsf{Undesired} \notin I$. Then, for all $a \in adom(I)$, $\mathsf{Accept}(a) \in I$ iff $a$ encodes a BCQ $q$ over $\mathcal{Q}$ (in the fixed way) such that $(D, q) \in O$.*

### 5.4 Generating universal model

Now it remains to show how to generate the universal model $U_O(D)$. Let $a$ be a number that encodes a BCQ $q$ such that $\mathsf{Accept}(a)$ is true in the intended instance. For all atoms $\alpha$ over $\mathcal{Q}$, we first test whether $\alpha$ appears in $q$. If the answer is yes we then copy $\alpha$ to the universal model. Since $U_O(D)$ is defined by a disjoint union of $[q]$, a renaming of variables in $q$ would be necessary. As variables are encoded by numbers, the renaming can be carried out by mapping a number to another number that is big enough. It is easy to see that the desired mapping can be defined by arithmetic relations.

With this assumption, we introduce two fresh relation symbols QVar and NewV where $\mathsf{QVar}(y, x)$ asserts that $y$ is an existential variable which appears in the BCQ encoded by $x$, and $\mathsf{NewV}(y, z, x)$ asserts that the variable $y$ which appears in the BCQ encoded by $x$ will be mapped to $z$. Now, we use them to define a new relation Name, where $\mathsf{Name}(y, z, x)$ means that $y$ will be replaced with $z$ in the copy of the BCQ encoded by $x$. Below are the DEDs that implement it.

$$\mathsf{BCQ}(x), \mathsf{QVar}(y, x), \mathsf{NewV}(y, z, x) \to \mathsf{Name}(y, z, x) \qquad (27)$$

$$\mathsf{BCQ}(x), \mathsf{DC}(y) \to \mathsf{Name}(y, y, x) \qquad (28)$$

where the second one means that we do not change constants.

Next, to copy all the atoms that appear in the BCQ to the intended universal model, we employ the following DEDs:

$$\mathsf{Accept}(x), \mathsf{HasQ}_i(\vec{y}, x), \mathsf{Name}(\vec{y}, \vec{z}, x) \to \mathsf{Q}_i(\vec{z}) \qquad (29)$$

where $\mathsf{Name}(\vec{y}, \vec{z}, x)$ denotes $\bigwedge_{1 \leq j \leq k} \mathsf{Name}(y_j, z_j, x)$ if $\vec{y} = y_1 \cdots y_k$, $\vec{z} = z_1 \cdots z_k$, and $k$ is the arity of $\mathsf{Q}_i$.

In addition, let us focus on an instance which defines a bad successor relation. To avoid changing the semantics, we have to force the instance to accept all BCQs, which can be done by defining the following DED for each $k$-ary $\mathsf{Q}_i$:

$$\mathsf{Undesired}, \mathsf{DC}(x_1), \ldots, \mathsf{DC}(x_k) \to \mathsf{Q}_i(x_1, \ldots, x_k) \qquad (30)$$

Finally, let $\Sigma_{um}$ denote the set of all DEDs defined here.

**Proposition 14.** *Let $D$ be a database over $\mathcal{D}$, and let $I$ be a minimal model of $D \cup \Sigma_{num} \cup \Sigma_{sim} \cup \Sigma_{um}$ such that $\mathsf{Undesired} \notin I$. Then $I|_\mathcal{Q}$ is isomorphic to $U_O(D)$.*

Now, by Propositions 11 and 14 we have Theorem 7.

## 6 Related work and discussions

Chandra and Harel [1980] proposed a language called QL and showed that it is expressively complete for all recursively enumerable database queries. Rudolph and Thomazo [2015] identified an elegant characterization for the expressiveness of TGDs, which asserts that TGDs define all the recursively enumerable database queries which are closed under homomorphisms. However, both languages, QL and TGDs, were treated as classical database query languages in their work. Our work focuses on the expressiveness of DEDs as a language for ontology-based query answering. As an immediate consequence, we have also established a characterization for the expressiveness of DEDs as a classical query language.

Gottlob *et al.* [2014a] proved that an ontology expressed by a finite set of TGDs and negative constraints (NCs) admits first-order rewritings if, and only if, it can be expressed by a finite set of TGDs and NCs that enjoys the bounded derivation depth property, which can be regarded as a relative expressive completeness w.r.t. a semantic class of existential rules. It would be interesting to know whether their characterization can be generalized to arbitrary OCQA ontologies. Zhang *et al.* [2015] showed that every TGD set with finite Skolem chase can be rewritten as a weakly-acyclic one, which is known as another relatively expressive completeness result.

It is also worth mentioning some other work related to ours. Gottlob *et al.* [2014b] showed that, as a classical query language, the weakly-guarded TGDs with stratified negations capture the class of queries decidable in EXPTIME. Arenas *et al.* [2014] proposed an interesting notion called *program expressive power*, and used it to compare the expressiveness of several ontological languages. Our framework for OCQA ontology can be regarded as a significant refinement of theirs.

The idea presented in this paper may be applied to other languages. For example, we can use it to prove that TGDs capture all recursively enumerable OCQA ontologies that are closed under database homomorphisms. The only additional difficulty is how to define a linear order by TGDs, which can be achieved by using some techniques from [Rudolph and

Thomazo, 2015; Gottlob *et al.*, 2014b]. Our work may shed a new light on identifying decidable languages for OCQA, and in particular on identifying an expressively complete language for first-order rewritable or tractable OCQA if it exists.

## Acknowledgements

We would like to thank the anonymous referees for their comments and suggestions on improving the paper. We are also grateful to Professor Moshe Y. Vardi for kindly pointing out an important related work. The first author's research was partially supported by the National Natural Science Foundation of China under grants 61173170, 61300222, 61370230, 61433006, 61572221 and U1401258.

# A  Appendix: Detailed Proofs

## A.1  Proof of Lemma 1

**Lemma 1.** *Let $q$ be a BCQ, and let $\Sigma$ and $\Gamma$ be two sets of DEDs such that $\Sigma$ is well-founded and possibly involves constants, no relation symbol from the relational schema of $\Sigma$ is intensional w.r.t. $\Gamma$, and all relation symbols from the relational schema of $q$ are intensional w.r.t. $\Gamma$. Then $\Sigma \cup \Gamma \models q$ iff $I \cup \Gamma \models q$ for all minimal models $I$ of $\Sigma$.*

*Proof.* The direction "only-if" is trivial. Thus, it remains to show the converse. Assume $I \cup \Gamma \models q$ for all minimal models $I$ of $\Sigma$. Now we want to prove $\Sigma \cup \Gamma \models q$. Let $J$ be an instance that satisfies $\Sigma \cup \Gamma$. Since $\Sigma$ is well-founded, there must exist a minimal model $I$ of $\Sigma$ such that $I \subseteq J$. Let $J_0 = I \cup J|_{\mathcal{R}}$ where $\mathcal{R}$ is the set of relation symbols that occur in the head of some DED in $\Gamma$. As relation symbols occurring in $I$ has no occurrence in the head of any DED in $\Gamma$, we know that $J_0$ is also a model of $\Gamma$. By the assumption, we obtain $J_0 \models q$. According to the definition of $J_0$, it must be true that $J \models q$, which implies $\Sigma \cup \Gamma \models q$ as desired. □

## A.2  Proof of Proposition 2

**Proposition 2.** *Let $\mathcal{R}$ be a relational schema; $\Phi$ be an abstract theory over $\mathcal{R}$; $D, D'$ be databases over $\mathcal{R}$; and $q, q'$ are boolean queries over $\mathcal{R}$. Then all the following hold:*

1. *If $D \cup \Phi \models q$ and $D \cup \Phi \models q'$ then $D \cup \Phi \models q \wedge q'$.*
2. *If $q \models q'$ and $D \cup \Phi \models q$ then $D \cup \Phi \models q'$.*
3. *If $D \Rightarrow_C D'$ and $D \cup \Phi \models q$ then $D' \cup \Phi \models q$, where $C$ denotes the set of constants that occur in $q$.*

*Proof.* Properties 1 and 2 are trivial. We only show the third one. Let us assume $D \Rightarrow_C D'$ and $D \cup \Phi \models q$. Our task is to show that $D' \cup \Phi \models q$. Let $A$ be a structure from $\Phi$ such that $A \models D'$. Then we have $D' \subseteq I(A)$. Since $D \Rightarrow_C D'$, there is an injective homomorphism $h : D \to D'$ such that $h(c) = c$ for all $c \in C$. Let $B$ be a structure over $\mathcal{R}$ such that

- $adom(D) \subseteq dom(B)$, and
- there exists an isomorphism $p$ from $A$ into $B$ such that $p(a) = b$ if $h(b) = a$ and that $p(c) = c$ for all $c \in C$.

It is easy to see that such a structure exists. As $\Phi$ is closed under isomorphisms, we know that $B \in \Phi$. It is also clear that $B \models D$. From $D \cup \Phi \models q$ we obtain that $B \models q$. By an induction on $q$, one can show $A \models q$ as desired. □

## A.3  Proof of Proposition 4

**Proposition 4.** *Let $\mathcal{D}, \mathcal{Q}$ be two disjoint relational schemas, and $O$ be a quasi-OCQA ontology over $(\mathcal{D}, \mathcal{Q})$. Then $O$ can be defined by a possibly infinite set of DEDs iff it is an OCQA ontology.*

*Proof.* The direction "only-if" follows from Proposition 2 immediately. So we only need to consider the converse.

Let $O$ be an OCQA ontology over $(\mathcal{D}, \mathcal{Q})$. Given any constant $a$, we introduce $v_a$ as a fresh variable. For each formula $\alpha$, let $\alpha^v$ denote the one obtained from $\alpha$ by substituting $v_a$ for $a$ if $a$ is a constant. Given any pair $(D, q) \in O$, let $\sigma_D^q$ be short for the DED $\varphi_D \to q^v \vee \psi_D$, where $\varphi_D$ is the conjunction of $\alpha^v$ for all facts $\alpha \in D$, and $\psi_D$ denotes the disjunction of equalities $v_a = v_b$ for all pairs of disjoint constants $a, b \in adom(D)$. Let $\Sigma_O$ be the set of DEDs $\sigma_D^q$ for all pairs $(D, q) \in O$. We want to show that $\Sigma_O$ defines $O$ over $(\mathcal{D}, \mathcal{Q})$, that is, $O = Sem(\Sigma_O, \mathcal{D}, \mathcal{Q})$.

Let $(D, q) \in O$. Then we have $\sigma_D^q \in \Sigma_O$. Let $\mathcal{R}$ be the relational schema of $\Sigma_O$. Let $I$ be an instance over $\mathcal{R}$ such that $D \subseteq I$ and $I \models \Sigma_O$. Let $s$ be a substitution that maps $v_a$ to $a$ for all $a \in adom(D)$. Then it is clear that $s(\varphi_D) \subseteq D \subseteq I$. Since $s(\psi_D)$ is obviously false, we conclude that $s(q^v) = q$ must be satisfied by $I$. This yields that $D \cup \Sigma_O \models q$, or equivalently, $(D, q) \in Sem(\Sigma_O, \mathcal{D}, \mathcal{Q})$ as desired.

Conversely, let $(D, q) \in Sem(\Sigma_O, \mathcal{D}, \mathcal{Q})$, which means that $D \cup \Sigma_O \models q$. Thus, for all $U \in Chase(D, \Sigma_O)$ there is a homomorphism, say $h$, from $[q]$ to $U$. Without loss of generality, we assume $q$ is of the form $q_1 \wedge \cdots \wedge q_k$, where each $q_i$ is a prime BCQ. By the definition of chase, for each $i$, there exist a BCQ $p_i$ and a database $D' \subseteq D$ such that $p_i \models q_i$ and $\sigma_{D'}^{p_i} \in \Sigma_O$. From the latter we know $(D', p_i) \in O$. Since $O$ is preserved under query implications, we infer $(D', q_i) \in O$. As $O$ is preserved under strict database homomorphisms, we obtain $(D, q_i) \in O$. By the query conjunction preservation of $O$, we conclude $(D, q) \in O$, which completes the proof. □

## A.4  Proof of Proposition 5

**Proposition 5.** *There is an OCQA ontology that is not definable by any first-order sentence.*

*Proof.* We consider an OCQA ontology that encodes the recurring domino problem. Let us first recall some notions. Every *domino system* is defined as a triple $(D, H, V)$ where $D$ is a finite set of dominos and $H, V \subseteq D \times D$ are two binary relations on $D$. Let $\mathbb{Z}$ be the set of integers. Given a domino system $S = (D, H, V)$, we call a function $\tau : \mathbb{Z} \times \mathbb{Z} \to D$ a *tiling of $S$ to the plane $\mathbb{Z} \times \mathbb{Z}$*, if for all integers $i, j$ we have

$$(\tau(i,j), \tau(i+1,j)) \in H \ \& \ (\tau(i,j), \tau(i,j+1)) \in V. \quad (31)$$

A tiling $\tau$ of $S$ to $\mathbb{Z} \times \mathbb{Z}$ is called *recurring w.r.t.* some domino $d \in D$ if there is an infinite number of pairs $(i, j)$ such that $\tau(i, j) = d$. It had been proved by Harel [1986] that deciding, given a domino system $S = (D, H, V)$ and a domino $d \in D$, whether there is a recurring tiling $\tau$ of $S$ to the plane $\mathbb{Z} \times \mathbb{Z}$, is $\Sigma_1^1$-complete. By definition, it is easy to see that, given two domino systems $S_1 = (D_1, H_1, V_1)$ and $S_2 = (D_2, H_2, V_2)$, if $H_1 \subseteq H_2$ and $V_1 \subseteq V_2$, then every recurring tiling of $S_1$ to $\mathbb{Z} \times \mathbb{Z}$ is also a recurring tiling of $S_2$ to $\mathbb{Z} \times \mathbb{Z}$.

Let $\mathcal{D} = \{\mathsf{H}, \mathsf{V}\}$ and $\mathcal{Q} = \{\mathsf{RecTiling}\}$, where $\mathsf{H}$ and $\mathsf{V}$ are binary relation symbols intended to define the binary relations $H$ and $V$, respectively, in a given domino system, and $\mathsf{RecTiling}$ is a nullary relation symbol that specifies whether the given domino system has a recurring tiling to the plane $\mathbb{Z} \times \mathbb{Z}$. It is easy to see that every domino system $S$ can be naturally encoded by a $\mathcal{D}$-database $D_S$. Let $O$ denote the set of ordered pairs $(D_S, \mathsf{RecTiling})$ for all domino systems $S$ that admits a recurring tiling to the plane $\mathbb{Z} \times \mathbb{Z}$. From the observation mentioned in the end of the last paragraph, one can easily verify that $O$ is indeed an OCQA ontology.

Towards a contradiction, we assume that $O$ is defined by a first-order sentence $\varphi$. Given a domino system $S$, to check whether $(D_S, \mathsf{RecTiling}) \in O$ or not, it is sufficient to check whether $D_S \cup \{\varphi\} \models \mathsf{RecTiling}$ or not. It is clear that the latter is recursively enumerable. However, this contradicts with the mentioned $\Sigma_1^1$-completeness, which is desired. □

## A.5 Proof of Proposition 9

**Proposition 9.** *There is a first-order rewritable OCQA ontology that cannot be defined by any finite set of DTGDs.*

This result can be proved by showing that every OCQA ontology defined by DTGDs is closed under database homomorphisms. Example 1 gives us an OCQA ontology that is not closed under database homomorphisms. It is easy to check that this OCQA ontology is first-order rewritable. So, it suffices to show the database homomorphism preservation for OCQA ontologies defined by DTGDs. Now let us present the property. Note that a simpler case (for TGDs) of this property was firstly observed by Rudolph and Thomazo [2015].

**Proposition 15.** *Let $\mathcal{R}$ be a relational schema, let $\Sigma$ and $q$ be a set of DTGDs and a BCQ over $\mathcal{R}$, respectively, let $C$ denote $const(q)$, and let $D, D'$ be two databases over $\mathcal{R}$ such that $D \to_C D'$ and $D \cup \Sigma \models q$. Then it holds that $D' \cup \Sigma \models q$.*

*Proof.* By a routine induction on the chase procedure. □

## A.6 Proof of Proposition 10

**Proposition 10.** *There is a first-order rewritable OCQA ontology that cannot be defined by any finite set of EDs.*

To prove this, the general idea is as follows. We first show that the complement of every OCQA ontology defined by EDs is closed under direct products. Then construct a OCQA ontology which does not enjoy the mentioned property.

Now we prove the inverse direct product preservation, first recall some notions. Let $\mathcal{R}$ be a relational schema, and $I$ and $J$ be two instances over $\mathcal{R}$. The *direct product* of $I$ and $J$, denoted $I \times J$, is a instance over $\mathcal{R}$ such that for all relation symbols $\mathsf{R} \in \mathcal{R}$ of arity $k$ for some $k \geq 0$, all $k$-tuple $\vec{a} \in adom(I)^k$ and all $k$-tuple $\vec{b} \in dom(J)^k$, we have

$$\mathsf{R}(\langle a_1, b_1\rangle, \ldots, \langle a_k, b_k\rangle) \in I \times J \quad (32)$$

iff both $\mathsf{R}(\vec{a}) \in I$ and $\mathsf{R}(\vec{b}) \in J$ holds, where $a_i$ (resp., $b_i$) denotes the $i$-th component of $\vec{a}$ (resp., $\vec{b}$).

A negative constraints (NCs) is a sentence of the form

$$\forall \vec{x}.(\phi(\vec{x}) \to \bot) \quad (33)$$

where $\phi(\vec{x})$ is a conjunction of relation atoms involving only terms from $\vec{x}$.

We first show the following property.

**Lemma 16.** *Every set $\Sigma$ of EDs and NCs is preserved under direct products. That is, for all instances $I$ and $J$ over the relational schema of $\Sigma$, $I \models \Sigma$ and $J \models \Sigma$ implies $I \times I' \models \Sigma$.*

*Proof.* By definition and a routine check. □

**Proposition 17.** *Let $\mathcal{R}$ be a relational schema, let $\Sigma$ and $q$ be a set of EDs and a BCQ over $\mathcal{R}$, respectively, let $C$ denote $const(q)$, and let $D, D'$ be two databases over $\mathcal{R}$ such that $D \cup \Sigma \not\models q$ and $D' \cup \Sigma \not\models q$. Then we have $D \times D' \cup \Sigma \not\models q$.*

*Proof.* Let $\Sigma_q$ denote the union of $\Sigma$ and $\{q \to \bot\}$. Then by Lemma 16, $\Sigma_q$ is preserved under direct products. It is also clear that, for any database $D_0$ over $\mathcal{R}$, $D_0 \cup \Sigma \not\models q$ iff $D_0 \cup \Sigma_q$ has a model, or equivalently, there is an instance $I_0 \supseteq D_0$ such that $I_0 \models \Sigma_q$. Let $I$ be an instance over $\mathcal{R}$ such that $I \supseteq D$ and $I \models \Sigma_q$, and let $I'$ be a similar instance for $D'$. Since $\Sigma_q$ is preserved under direct products, we have $I \times I' \models \Sigma_q$. On the other hand, by definition, it is easy to see that $D \times D' \subseteq I \times I'$. Note that we have $D \subseteq I$ and $D' \subseteq I'$. Combining this with the previous conclusion, we know that $D \times D' \cup \Sigma \not\models q$, which completes the proof. □

Next, let us present a counterexample.

Let $\mathcal{D} = \{\mathsf{A}\}$ and $\mathcal{Q} = \{\mathsf{Goal}\}$, where A is a binary relation symbol, and Goal a nullary relation symbol. Let $\Sigma$ consist of a single DED as follows:

$$\bigwedge_{0 \leq i,j < 4} \mathsf{A}(x_i, x_j) \to \mathsf{Goal} \vee \bigvee_{0 \leq i < j < 4} x_i = x_j. \quad (34)$$

Let $O$ denote the OCQA ontology $Sem(\Sigma, \mathcal{D}, \mathcal{Q})$. It is easy to show that $O$ is first-order rewritable. Note that the only BCQ over $\mathcal{Q}$ is Goal, and that $D \cup \Sigma \models \mathsf{Goal}$ iff

$$D \models \exists x_0 \cdots \exists x_3. \left[\bigwedge_{0 \leq i,j < 4} \mathsf{A}(x_i, x_j) \wedge \bigwedge_{0 \leq i < j < 4} \neg x_i = x_j\right] \quad (35)$$

for all databases $D$ over $\mathcal{D}$.

Let $D'$ denote a database over $\mathcal{R}$ as follows:

$$\{\mathsf{A}(a, a), \mathsf{A}(a, b), \mathsf{A}(b, a), \mathsf{A}(b, b)\} \quad (36)$$

where $a, b$ are distinct constants. It is clear that $D' \cup \Sigma \models \mathsf{Goal}$ does not hold. However, it is easy to see that $D' \times D' \cup \Sigma \models \mathsf{Goal}$. This means that $O = Sem(\Sigma, \mathcal{D}, \mathcal{Q})$ is an OCQA ontology that does not satisfy the property presented in Proposition 17, which completes the proof of Proposition 10.